\documentclass[11pt,a4paper]{article}
\usepackage[hyperref]{emnlp-ijcnlp-2019}
\usepackage{times}
\usepackage{latexsym}
\usepackage{amsfonts}
\usepackage{url}
\usepackage{bm} 
\usepackage{graphicx}
\usepackage{microtype}
\usepackage{subfigure}
\usepackage{booktabs} 

\usepackage{amsmath}
\usepackage{verbatim}
\usepackage{amsfonts}
\usepackage{rotating}
\usepackage{multirow}
\usepackage{colortbl}
\usepackage{fancyhdr}
\aclfinalcopy 

\title{Different Absorption from the Same Sharing: Sifted Multi-task Learning for Fake News Detection}

\author{Lianwei Wu, Yuan Rao, Haolin Jin, Ambreen Nazir, Ling Sun \\
  Lab of Social Intelligence and Complexity Data Processing \\ School of Software Engineering, Xi'an Jiaotong University,  Xi'an, 710049, China \\
  {\tt \{stayhungry,jinhaolin,ambreen.nazir,sunling\}@stu.xjtu.edu.cn} \\ {\tt raoyuan@mail.xjtu.edu.cn} \\}

\date{}

\begin{document}
\maketitle
\begin{abstract}
  Recently, neural networks based on multi-task learning have achieved promising performance on fake news detection, which focus on learning shared features among tasks as complementary features to serve different tasks. However, in most of the existing approaches, the shared features are completely assigned to different tasks without selection, which may lead to some useless and even adverse features integrated into specific tasks. In this paper, we design a sifted multi-task learning method with a selected sharing layer for fake news detection. The selected sharing layer adopts gate mechanism and attention mechanism to filter and select shared feature flows between tasks. Experiments on two public and widely used competition datasets, i.e. RumourEval and PHEME, demonstrate that our proposed method achieves the state-of-the-art performance and boosts the F1-score by more than 0.87\%, 1.31\%, respectively.
\end{abstract}

\section{Introduction}
In recent years, the proliferation of fake news with various content, high-speed spreading, and extensive influence has become an increasingly alarming issue. A concrete instance\footnote{http://business.time.com/2013/04/24/how-does-one-fake-tweet-cause-a-stock-market-crash/} was cited by Time Magazine in 2013 when a false announcement of Barack Obama's injury in a White House explosion ``wiped off 130 Billion US Dollars in stock value in a matter of seconds". Other examples, an analysis of the US Presidential Election in 2016 \cite{allcott2017social} revealed that fake news was widely shared during the three months prior to the election with 30 million total Facebook shares of 115 known pro-Trump fake stories and 7.6 million of 41 known pro-Clinton fake stories. Therefore, automatically detecting fake news has attracted significant research attention in both industries and academia.

\begin{figure}
\centering
\includegraphics[width=0.35\textwidth]{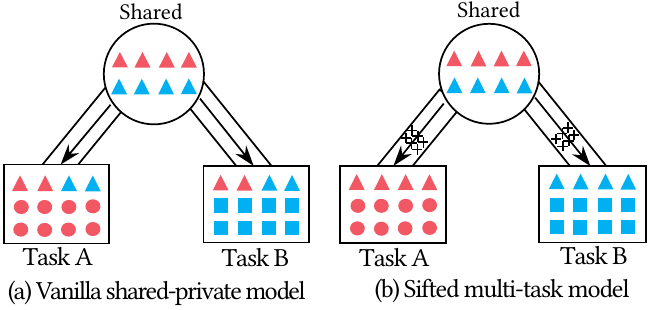}
\caption{Two schemes for sharing features among tasks. Red circles and blue boxes represent the task-specific features, while the red and blue triangles mean shared features that benefit Task A and Task B, respectively.}
\label{Fig1}
\end{figure}

Most existing methods devise deep neural networks to capture credibility features for fake news detection. Some methods provide in-depth analysis of text features, e.g., linguistic \cite{conroy2015automatic}, semantic \cite{yang2012automatic}, emotional \cite{wang2015unsupervised}, stylistic \cite{potthast2017stylometric}, etc. On this basis, some work additionally extracts social context features (a.k.a. meta-data features) as credibility features, including source-based \cite{castillo2011information}, user-centered \cite{long2017fake}, post-based \cite{wang2017liar} and network-based \cite{ruchansky2017csi}, etc. These methods have attained a certain level of success. Additionally, recent researches \cite{thorne2017fake,dungs2018can} find that doubtful and opposing voices against fake news are always triggered along with its propagation. Fake news tends to provoke controversies compared to real news \cite{mendoza2010twitter,zubiaga2016analysing}. Therefore, stance analysis of these controversies can serve as valuable credibility features for fake news detection.

There is an effective and novel way to improve the performance of fake news detection combined with stance analysis, which is to build multi-task learning models to jointly train both tasks \cite{ma2018detect,kochkina2018all,li2018end}. These approaches model information sharing and representation reinforcement between the two tasks, which expands valuable features for their respective tasks. However, prominent drawback to these methods and even typical multi-task learning methods, like the shared-private model, is that the shared features in the shared layer are equally sent to their respective tasks without filtering, which causes that some useless and even adverse features are mixed in different tasks, as shown in Figure \ref{Fig1}(a). By that the network would be confused by these features, interfering effective sharing, and even mislead the predictions.

To address the above problems, we design a sifted multi-task learning model with filtering mechanism (Figure \ref{Fig1}(b)) to detect fake news by joining stance detection task. Specifically, we introduce a selected sharing layer into each task after the shared layer of the model for filtering shared features. The selected sharing layer composes of two cells: gated sharing cell for discarding useless features and attention sharing cell for focusing on features that are conducive to their respective tasks. Besides, to better capture long-range dependencies and improve the parallelism of the model, we apply transformer encoder module \cite{vaswani2017attention} to our model for encoding input representations of both tasks. Experimental results reveal that the proposed model outperforms the compared methods and gains new benchmarks.

In summary, the contributions of this paper are as follows:

\begin{itemize}
    \item We explore a selected sharing layer relying on gate mechanism and attention mechanism, which can selectively capture valuable shared features between tasks of fake news detection and stance detection for respective tasks.
    \item The transformer encoder is introduced into our model for encoding inputs of both tasks, which enhances the performance of our method by taking advantages of its long-range dependencies and parallelism.
    \item Experiments on two public, widely used fake news datasets demonstrate that our method significantly outperforms previous state-of-the-art methods.
\end{itemize}

\section{Related Work}

\textbf{Fake News Detection } Exist studies for fake news detection can be roughly summarized into two categories. The first category is to extract or construct comprehensive and complex features with manual ways \cite{castillo2011information,ruchansky2017csi,flintham2018falling}. The second category is to automatically capture deep features based on neural networks. There are two ways in this category. One is to capture linguistic features from text content, such as semantic \cite{wang2017liar,wu2018false}, writing styles \cite{potthast2017stylometric}, and textual entailments \cite{oshikawa2018survey}. The other is to focus on gaining effective features from the organic integration of text and user interactions \cite{qian2018neural,wu2019multi}. User interactions include users' behaviours, profiles, and networks between users. In this work, following the second way, we automatically learn representations of text and stance information from response and forwarding (users' behaviour) based on multi-task learning for fake news detection.

\textbf{Stance Detection } The researches \cite{lukasik2016hawkes,zubiaga2016stance} demonstrate that the stance detected from fake news can serve as an effective credibility indicator to improve the performance of fake news detection. The common way of stance detection in rumors is to catch deep semantics from text content based on neural networks\cite{mohtarami2018automatic}. For instance, Kochkina et al.\cite{kochkina2017turing} project branch-nested LSTM model to encode text of each tweet considering the features and labels of the predicted tweets for stance detection, which reflects the best performance in RumourEval dataset. In this work, we utilize transformer encoder to acquire semantics from responses and forwarding of fake news for stance detection.

\begin{figure*}
\centering
\includegraphics[width=0.60\textwidth]{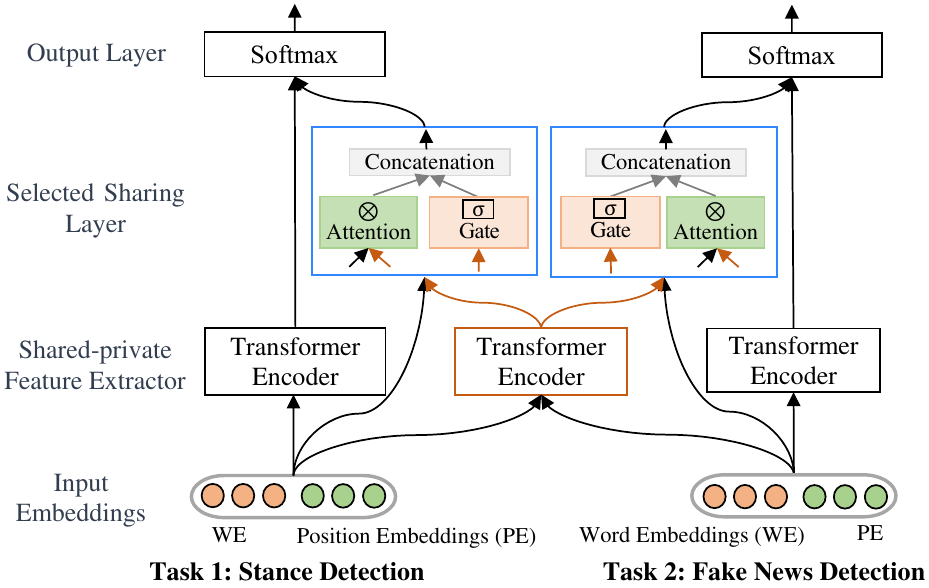}
\caption{The architecture of the sifted multi-task learning method based on shared-private model. In particular, the two blue boxes represent selected sharing layers of stance detection and fake news detection and the red box denotes shared layer between tasks.}
\label{Fig2}
\end{figure*}

\textbf{Multi-task Learning }  A collection of improved models \cite{chen2018multinomial,chen2018zero,liu2019augmented} are developed based on multi-task learning.
Especially, shared-private model, as a popular multi-task learning model, divides the features of different tasks into private and shared spaces, where shared features, i.e., task-irrelevant features in shared space,  as supplementary features are used for different tasks. Nevertheless, the shared space usually mixes some task-relevant features, which makes the learning of different tasks introduce noise. To address this issue, Liu et al. \cite{liu2017adversarial} explore an adversarial shared-private model to alleviate the shared and private latent feature spaces from interfering with each other. However, these models transmit all shared features in the shared layer to related tasks without distillation, which disturb specific tasks due to some useless and even harmful shared features. How to solve this drawback is the main challenge of this work.

\section{Method}

We propose a novel sifted multi-task learning method on the ground of shared-private model to jointly train the tasks of stance detection and fake news detection, filter original outputs of shared layer by a selected sharing layer. Our model consists of a 4-level hierarchical structure, as shown in Figure \ref{Fig2}. Next, we will describe each level of our proposed model in detail.

\subsection{Input Embeddings}

In our notation, a sentence of length $l$ tokens is indicated as ${\rm \textbf{X}}=\{x_1, x_2, ... ,x_l\}$. Each token is concatenated by word embeddings and position embeddings. Word embeddings $w_i$ of token $x_i$ are a $d_w$-dimensional vector obtained by pre-trained Word2Vec model \cite{mikolov2013distributed}, i.e., $w_i \in \mathbb{R}^{d_w}$. Position embeddings refer to vectorization representations of position information of words in a sentence. We employ one-hot encoding to represent position embeddings $p_i$ of token $x_i$, where $p_i \in \mathbb{R}^{d_p}$, $d_p$ is the positional embedding dimension. Therefore, the embeddings of a sentence are represented as $ {\rm \textbf{E}}=\{[w_1;p_1 ], [w_2;p_2], ..., [w_l;p_l]\}, {\rm \textbf{E}}\in \mathbb{R}^{l \times (d_p+d_w)}$. In particular, we adopt one-hot encoding to embed positions of tokens,  rather than sinusoidal position encoding recommended in BERT model \cite{devlin2018bert}. The reason is that our experiments show that compared with one-hot encoding, sinusoidal position encoding not only increases the complexity of models but also performs poorly on relatively small datasets.

\subsection{Shared-private Feature Extractor}
\label{subsection3.2}

Shared-private feature extractor is mainly used for extracting shared features and private features among different tasks. In this paper, we apply the encoder module of transformer \cite{vaswani2017attention} (henceforth, transformer encoder) to the shared-private extractor of our model. Specially, we employ two transformer encoders to encode the input embeddings of the two tasks as their respective private features. A transformer encoder is used to encode simultaneously the input embeddings of the two tasks as shared features of both tasks. This process is illustrated by the shared-private layer of Figure \ref{Fig2}. The red box in the middle denotes the extraction of shared features and the left and right boxes represent the extraction of private features of two tasks. Next, we take the extraction of the private feature of fake news detection as an example to elaborate on the process of transformer encoder.

The kernel of transformer encoder is the scaled dot-product attention, which is a special case of attention mechanism. It can be precisely described as follows:
\begin{eqnarray}\label{eq1}
    {\rm Attention}({\rm \textbf{Q}}, {\rm \textbf{K}}, {\rm \textbf{V}})={\rm softmax}(\frac{ {\rm \textbf{Q}}{\rm \textbf{K}}^T}{\sqrt{{\rm \textbf{d}}}}){\rm \textbf{V}}
\end{eqnarray}
where ${\rm \textbf{Q}} \in \mathbb{R}^{l \times (d_p+d_w)}$, ${\rm \textbf{K}} \in \mathbb{R}^{l \times (d_p+d_w)}$, and ${\rm \textbf{V}} \in \mathbb{R}^{l \times (d_p+d_w)}$ are query matrix, key matrix, and value matrix, respectively. In our setting, the query ${\rm \textbf{Q}}$ stems from the inputs itself, i.e., ${\rm \textbf{Q}}={\rm \textbf{K}}={\rm \textbf{V}}={\rm \textbf{E}}$.

To explore the high parallelizability of attention, transformer encoder designs a multi-head attention mechanism based on the scaled dot-product attention. More concretely, multi-head attention first linearly projects the queries, keys and values $h$ times by using different linear projections. Then $h$ projections perform the scaled dot-product attention in parallel. Finally, these results of attention are concatenated and once again projected to get the new representation. Formally, the multi-head attention can be formulated as follows:
\begin{equation}\label{eq2}
    head_i={\rm Attention}({\rm \textbf{Q}}{\rm \textbf{W}}^Q_i, {\rm \textbf{K}}{\rm \textbf{W}}^K_i, {\rm \textbf{V}}{\rm \textbf{W}}^V_i)
\setlength{\belowdisplayskip}{0pt}
\end{equation}
\begin{equation}\label{eq3}
\setlength{\abovedisplayskip}{0pt}
  \begin{aligned}
    {\rm \textbf{H}} &={\rm MultiHead}({\rm \textbf{Q}}, {\rm \textbf{K}}, {\rm \textbf{V}})\\
    &={\rm Concat}(head_1, head_2, ..., head_h){\rm \textbf{W}}^o
  \end{aligned}
\end{equation}
where ${\rm \textbf{W}}_i^Q \in \mathbb{R}^{(d_p+d_w) \times d_k}$, ${\rm \textbf{W}}_i^K \in \mathbb{R}^{(d_p+d_w) \times d_k}$, ${\rm \textbf{W}}_i^V \in \mathbb{R}^{(d_p+d_w) \times d_k}$ are trainable projection parameters. $d_k$ is $(d_p+d_w)/h$, $h$ is the number of heads. In Eq.(\ref{eq3}), ${\rm \textbf{W}}^o \in \mathbb{R}^{(d_p+d_w) \times (d_p+d_w)}$ is also trainable parameter.

\subsection{Selected Sharing Layer}

\begin{figure}
\centering
\includegraphics[width=0.3\textwidth]{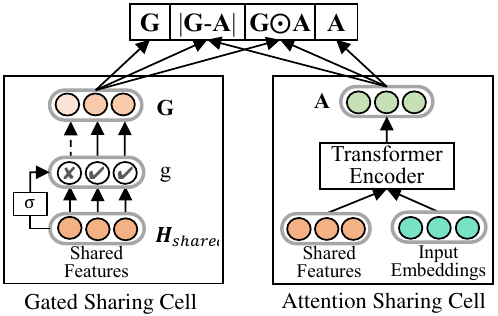}
\caption{The details of selected sharing layer.}
\label{Fig3}
\end{figure}

In order to select valuable and appropriate shared features for different tasks, we design a selected sharing layer following the shared layer. The selected sharing layer consists of two cells: gated sharing cell for filtering useless features and attention sharing cell for focusing on valuable shared features for specific tasks. The description of this layer is depicted in Figure \ref{Fig2} and Figure \ref{Fig3}. In the following, we introduce two cells in details.

\textbf{Gated Sharing Cell } Inspired by forgotten gate mechanism of LSTM \cite{hochreiter1997long} and GRU \cite{chung2014empirical}, we design a single gated cell to filter useless shared features from shared layer. There are two reasons why we adopt single-gate mechanism. One is that transformer encoder in shared layer can efficiently capture the features of long-range dependencies. The features do not need to capture repeatedly by multiple complex gate mechanisms of LSTM and GRU. The other is that single-gate mechanism is more convenient for training \cite{srivastava2015highway}. Formally, the gated sharing cell can be expressed as follows:
\begin{equation}\label{eq4}
    {\rm \textbf{g}}_{fake} = \sigma ({\rm \textbf{W}}_{fake} \cdot {\rm \textbf{H}}_{shared} + {\rm \textbf{b}}_{fake})
\end{equation}
where ${\rm \textbf{H}}_{shared}\! \in \! \mathbb{R}^{1 \times l(d_p+d_w)}$ \! denotes the outputs of shared layer upstream, ${\rm \textbf{W}}_{fake} \in \mathbb{R}^{l(d_p+d_w) \times l(d_p+d_w)}$ and ${\rm \textbf{b}}_{fake} \in \mathbb{R}^{1 \times l(d_p+d_w)}$ are trainable parameters. $\sigma$ is a non-linear activation - sigmoid, which makes final choices for retaining and discarding features in shared layer.

Then the shared features after filtering via gated sharing cell ${\rm \textbf{g}}_{fake}$ for the task of fake news detection are represented as:
\begin{equation}\label{eq5}
    {\rm \textbf{G}}_{fake}={\rm \textbf{g}}_{fake} \odot {\rm \textbf{H}}_{shared}
\end{equation}
where $\odot$ denotes element-wise multiplication.

Similarly, for the auxiliary task - the task of stance detection, filtering process in the gated sharing cell is the same as the task of fake news detection, so we do not reiterate them here.

\textbf{Attention Sharing Cell } To focus on helpful shared features that are beneficial to specific tasks from upstream shared layer, we devise an attention sharing cell based on attention mechanism. Specifically, this cell utilizes input embeddings of the specific task to weight shared features for paying more attention to helpful features. The inputs of this cell include two matrixes: the input embeddings of the specific task and the shared features of both tasks. The basic attention architecture of this cell, the same as shared-private feature extractor, also adopts transformer encoder (the details in subsection \ref{subsection3.2}). However, in this architecture, query matrix and key matrix are not projections of the same matrix, i.e., query matrix ${\rm \textbf{E}}_{fake}$ is the input embeddings of fake news detection task, and key matrix ${\rm \textbf{K}}_{shared}$ and value matrix ${\rm \textbf{V}}_{shared}$ are the projections of shared features ${\rm \textbf{H}}_{shared}$. Formally, the attention sharing cell can be formalized as follows:
\begin{equation}\label{eq6}
  \begin{aligned}
    head_i &={\rm Attention}( \\
    &{\rm \textbf{E}}_{fake}{\rm \textbf{W}}^Q_i, {\rm \textbf{K}}_{shared}{\rm \textbf{W}}^K_i, {\rm \textbf{V}}_{shared}{\rm \textbf{W}}^V_i)
  \end{aligned}
\setlength{\belowdisplayskip}{0pt}
\end{equation}
\begin{equation}\label{eq7}
\setlength{\abovedisplayskip}{0pt}
  \begin{aligned}
    {\rm \textbf{A}}_{fake} &={\rm MultiHead}({\rm \textbf{H}}_{fake}, {\rm \textbf{K}}_{shared}, {\rm \textbf{V}}_{shared})\\
    &={\rm Concat}(head_1, head_2, ..., head_h){\rm \textbf{W}}^o
  \end{aligned}
\end{equation}
where the dimensions of ${\rm \textbf{E}}_{fake}$, ${\rm \textbf{K}}_{shared}$, and ${\rm \textbf{V}}_{shared}$ are all $\mathbb{R}^{l\times (d_p+d_w)}$. The dimensions of remaining parameters in Eqs.(\ref{eq6}, \ref{eq7}) are the same as in Eqs.(\ref{eq2}, \ref{eq3}). Moreover, in order to guarantee the diversity of focused shared features, the number of heads $h$ should not be set too large. Experiments show that our method performs the best performance when $h$ is equal to 2.

\textbf{Integration of the Two Cells } We first convert the output of the two cells to vectors ${\rm \textbf{G}}$ and ${\rm \textbf{A}}$, respectively, and then integrate the vectors in full by the absolute difference and element-wise product \cite{mou2016natural}.
\begin{equation}\label{eq8}
    {\rm \textbf{SSL}}=[{\rm \textbf{G}};|{\rm \textbf{G}}-{\rm \textbf{A}}|;{\rm \textbf{G}} \odot {\rm \textbf{A}};{\rm \textbf{A}}]
\end{equation}
where $\odot$ denotes element-wise multiplication and $;$ denotes concatenation.

\subsection{The Output Layer}

As the last layer, softmax functions are applied to achieve the classification of different tasks, which emits the prediction of probability distribution for the specific task $i$.
\begin{equation}\label{eq9}
    \hat{{\rm \textbf{y}}}_i={\rm softmax}({\rm \textbf{W}}_i{\rm \textbf{F}}_i+{\rm \textbf{b}}_i)
\setlength{\belowdisplayskip}{0pt}
\end{equation}
\begin{equation}\label{eq10}
\setlength{\abovedisplayskip}{0pt}
    {\rm \textbf{F}}_i = [{\rm \textbf{H}}_i; {\rm \textbf{SSL}}_i]
\end{equation}
where $\hat{{\rm \textbf{y}}}_i$ is the predictive result, ${\rm \textbf{F}}_i$ is the concatenation of private features ${\rm \textbf{H}}_i$ of task $i$ and the outputs ${\rm \textbf{SSL}}_i$ of selected sharing layer for task $i$. ${\rm \textbf{W}}_i$ and ${\rm \textbf{b}}_i$ are trainable parameters.

Given the prediction of all tasks, a global loss function forces the model to minimize the cross-entropy of prediction and true distribution for all the tasks:
\begin{equation}\label{eq11}
    \mathcal{L} = \sum_{i=1}^{N} \lambda_i L(\hat{{\rm \textbf{y}}}_i, {\rm \textbf{y}}_i)
\setlength{\belowdisplayskip}{0pt}
\end{equation}
\begin{equation}\label{eq12}
\setlength{\abovedisplayskip}{0pt}
    L(\hat{{\rm \textbf{y}}}_i, {\rm \textbf{y}}_i) = {\rm \textbf{y}}_i log \hat{{\rm \textbf{y}}}_i + (1-{\rm \textbf{y}}_i) log(1-\hat{{\rm \textbf{y}}}_i)
\end{equation}
where $\lambda_i$ is the weight for the task $i$, and $N$ is the number of tasks. In this paper, $N=2$, and we give more weight $\lambda$ to the task of fake news detection.

\begin{table*}
\centering
\small
\begin{tabular}{|l|c|c|c|c|c|c|c|c|c|}
\hline
    Datasets & Threads & Tweets & True & False & Unverified & Support & Deny & Query & Comment\\\hline
    RumourEval & 325 & 5,568 & 145 & 74 & 106 & 1,004 & 415 & 464 & 3,685 \\\hline
    PHEME & 6,425 & 105,354 & 1,067 & 638 & 697 & 891 & 335 & 353 & 2,855 \\
\hline
\end{tabular}
\caption{Statistics of the two datasets.}
\label{Tab1}
\end{table*}

\section{Experiments}

\subsection{Datasets and Evaluation Metrics}

We use two public datasets for fake news detection and stance detection, i.e., RumourEval \cite{derczynski2017semeval} and PHEME \cite{zubiaga2016analysing}. We introduce both the datasets in details from three aspects: content, labels, and distribution.

\textbf{Content.} Both datasets contain Twitter conversation threads associated with different newsworthy events including the Ferguson unrest, the shooting at Charlie Hebdo, etc. A conversation thread consists of a tweet making a true and false claim, and a series of replies. \textbf{Labels.} Both datasets have the same labels on fake news detection and stance detection. Fake news is labeled as true, false, and unverified. Because we focus on classifying true and false tweets, we filter the unverified tweets. Stance of tweets is annotated as support, deny, query, and comment. \textbf{Distribution.} RumourEval contains 325 Twitter threads discussing rumours and PHEME includes 6,425 Twitter threads. Threads, tweets, and class distribution of the two datasets are shown in Table \ref{Tab1}.

In consideration of the imbalance label distributions, in addition to accuracy (A) metric, we add Precision (P), Recall (R) and F1-score (F1) as complementary evaluation metrics for tasks. We hold out 10\% of the instances in each dataset for model tuning, and the rest of the instances are performed 5-fold cross-validation throughout all experiments.

\subsection{Settings}

\textbf{Pre-processing } - Processing useless and inappropriate information in text: (1) removing nonalphabetic characters; (2) removing website links of text content; (3) converting all words to lower case and tokenize texts.

\textbf{Parameters } - hyper-parameters configurations of our model: for each task, we strictly turn all the hyper-parameters on the validation dataset, and we achieve the best performance via a small grid search. The sizes of word embeddings and position embeddings are set to 200 and 100. In transformer encoder, attention heads and blocks are set to 6 and 2 respectively, and the dropout of multi-head attention is set to 0.7. Moreover, the minibatch size is 64; the initial learning rate is set to 0.001, the dropout rate to 0.3, and $\lambda$ to 0.6 for fake news detection.

\begin{table*}
\centering
\small
\begin{tabular}{|l|l|c|c|c|c|c|c|c|c|}
\hline
    Dataset & Measure & SVM & CNN & TE & DeClarE & MTL-LSTM & TRNN & Bayesian-DL & Ours \\\hline
    \multirow{4}*{RumourEval} & A(\%) & 71.42 & 61.90 & 66.67 & 66.67 & 66.67 & 76.19 & 80.95 & \textbf{81.48} \\
    & P(\%) & 66.67 & 54.54 & 60.00 & 58.33 & 57.14 & 70.00 & \textbf{77.78} & 72.24 \\
    & R(\%) & 66.67 & 66.67 & 66.67 & 77.78 & \textbf{88.89} & 77.78 & 77.78 & 86.31 \\
    & F1(\%) & 66.67 & 59.88 & 63.15 & 66.67 & 69.57 & 73.68 & 77.78 & \textbf{78.65} \\\hline
    \multirow{4}*{PHEME} & A(\%) & 72.18 & 59.23 & 65.22 & 67.87 & 74.94 & 78.65 & 80.33 & \textbf{81.27} \\
    & P(\%) & \textbf{78.80} & 56.14 & 63.05 & 64.68 & 68.77 & 77.11 & 78.29 & 73.41 \\
    & R(\%) & 75.75 & 64.64 & 64.64 & 71.21 & 87.87 & 78.28 & 79.29 & \textbf{88.10} \\
    & F1(\%) & 72.10 & 60.09 & 63.83 & 67.89 & 77.15 & 77.69 & 78.78 & \textbf{80.09} \\
\hline
\end{tabular}
\caption{Performance comparison of our sifted multi-task learning method against the baselines.}
\label{Tab2}
\end{table*}

\subsection{Performance Evaluation}

\subsubsection{Baselines}

\textbf{SVM } A Support Vector Machines model in \cite{derczynski2017semeval} detects misinformation relying on manually extracted features.

\textbf{CNN } A Convolutional Neural Network model \cite{chen2017ikm} employs pre-trained word embeddings based on Word2Vec as input embeddings to capture features similar to n-grams.

\textbf{TE } Tensor Embeddings \cite{guacho2018semi} leverages tensor decomposition to derive concise claim embeddings, which are used to create a claim-by-claim graph for label propagation.

\textbf{DeClarE } Evidence-Aware Deep Learning \cite{popat2018declare} encodes claims and articles by Bi-LSTM and focuses on each other based on attention mechanism, and then concatenates claim source and article source information.

\textbf{MTL-LSTM } A multi-task learning model based on LSTM networks \cite{kochkina2018all} trains jointly the tasks of veracity classification, rumor detection, and stance detection.

\textbf{TRNN } Tree-structured RNN \cite{ma2018rumor} is a bottom-up and a top-down tree-structured model based on recursive neural networks.

\textbf{Bayesian-DL } Bayesian Deep Learning model \cite{zhang2019reply} first adopts Bayesian to represent both the prediction and uncertainty of claim and then encodes replies based on LSTM to update and generate a posterior representations.

\begin{table*}
\centering
\small
\begin{tabular}{|l|c|c|c|c|c|c|c|c|}
\hline
     & \multicolumn{4}{c|}{RumourEval} & \multicolumn{4}{c|}{PHEME} \\\hline
     & A(\%) & P(\%) & R(\%) & F1(\%) & A(\%) & P(\%) & R(\%) & F1(\%) \\\hline
    Single-task & 62.86 & 54.21 & 65.43 & 59.29 & 60.94 & 55.56 & 64.53 & 58.57 \\\hline
    MT-lstm & 65.90 & 56.61 & 85.60 & 68.15 & 71.61 & 66.24 & 85.31 & 75.29 \\
    MT-trans & 71.67 & 58.43 & 78.78 & 67.10 & 75.57 & 65.73 & 78.94 & 71.73 \\
    MT-trans-G & 74.89 & 64.32 & 81.68 & 71.97 & 76.24 & 67.32 & 83.56 & 74.57 \\
    MT-trans-A & 79.22 & 68.97 & 84.86 & 76.10 & 79.54 & 70.90 & 86.71 & 78.01 \\
    MT-trans-G-A & 82.10 & 72.24 & 86.31 & 78.65 & 81.27 & 73.41 & 88.10 & 80.09 \\
\hline
\end{tabular}
\caption{Ablation analysis of the sifted multi-task learning method.}
\label{Tab3}
\end{table*}

\subsubsection{Compared with State-of-the-art Methods}
We perform experiments on RumourEval and PHEME datasets to evaluate the performance of our method and the baselines. The experimental results are shown in Table \ref{Tab2}. We gain the following observations:

\begin{itemize}
\item On the whole, most well-designed deep learning methods, such as ours, Bayesian-DL, and TRNN, outperform feature engineering-based methods, like SVM. This illustrates that deep learning methods can represent better intrinsic semantics of claims and replies.
\item In terms of recall (R), our method and MTL-LSTM, both based on multi-task learning, achieve more competitive performances than other baselines, which presents that sufficient features are shared for each other among multiple tasks. Furthermore, our method reflects a more noticeable performance boost than MTL-LSTM on both datasets, which extrapolates that our method earns more valuable shared features.
\item Although our method shows relatively low performance in terms of precision (P) and recall (R) compared with some specific models, our method achieves the state-of-the-art performance in terms of accuracy (A) and F1-score (F1) on both datasets. Taking into account the tradeoff among different performance measures, this reveals the effectiveness of our method in the task of fake news detection.

\end{itemize}
\begin{figure*}
\centering
\includegraphics[width=0.8\textwidth]{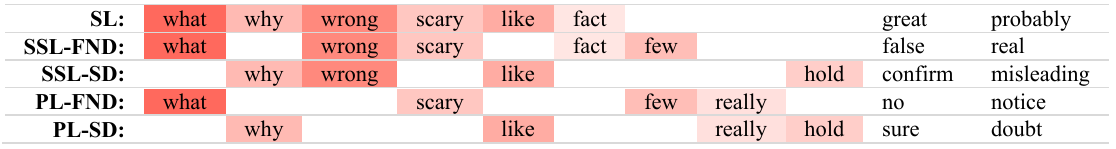}
\caption{Typical tokens obtained by different layers of the sifted multi-task learning method. In our proposed method, typical tokens are captured by shared layer (SL), selected sharing layer for fake news detection (SSL-FND), selected sharing layer for stance detection (SSL-SD), private layer for fake news detection (PL-FND), and private layer for stance detection (PL-SD) respectively. A column of the same color represents the distribution of one token in different layers, while the last two columns denote unique tokens captured by different layers.}
\label{Fig4}
\end{figure*}
\subsection{Discussions}

\subsubsection{Model Ablation}

To evaluate the effectiveness of different components in our method, we ablate our method into several simplified models and compare their performance against related methods. The details of these methods are described as follows:

\textbf{Single-task } Single-task is a model with transformer encoder as the encoder layer of the model for fake news detection.

\textbf{MT-lstm } The tasks of fake news detection and stance detection are integrated into a shared-private model and the encoder of the model is achieved by LSTM.

\textbf{MT-trans } The only difference between MT-trans and MT-lstm is that encoder of MT-trans is composed of transformer encoder.

\textbf{MT-trans-G } On the basis of MT-trans, MT-trans-G adds gated sharing cell behind the shared layer of MT-trans to filter shared features.

\textbf{MT-trans-A } Unlike MT-trans-G, MT-trans-A replaces gated sharing cell with attention sharing cell for selecting shared features.

\textbf{MT-trans-G-A } Gated sharing cell and attention sharing cell are organically combined as selected sharing layer behind the shared layer of MT-trans, called MT-trans-G-A.

Table \ref{Tab3} provides the experimental results of these methods on RumourEval and PHEME datasets. We have the following observations:

\begin{itemize}
    \item Effectiveness of multi-task learning. MT-trans boosts about 9\% and 15\% performance improvements in accuracy on both datasets compared with Single-task, which indicates that the multi-task learning method is effective to detect fake news.
    \item Effectiveness of transformer encoder. Compared with MT-lstm, MT-trans obtains more excellent performance, which explains that transformer encoder has better encoding ability than LSTM for news text on social media.
    \item Effectiveness of the selected sharing layer. Analysis of the results of the comparison with MT-trans, MT-trans-G, MT-Trans-A, and MT-trans-G-A shows that MT-trans-G-A ensures optimal performance with the help of the selected sharing layer of the model, which confirms the reasonability of selectively sharing different features for different tasks.
\end{itemize}

\subsubsection{Error Analysis}

Although the sifted multi-task learning method outperforms previous state-of-the-art methods on two datasets (From Table \ref{Tab2}), we observe that the proposed method achieves more remarkable performance boosts on PHEME than on RumourEval. There are two reasons for our analysis according to Table \ref{Tab1} and Table \ref{Tab2}. One is that the number of training examples in RumourEval (including 5,568 tweets) is relatively limited as compared with PHEME (including 105,354 tweets), which is not enough to train deep neural networks.
Another is that PHEME includes more threads (6,425 threads) than RumourEval (325 threads) so that PHEME can offer more rich credibility features to our proposed method.
\begin{figure*}
\centering
\subfigure[Obtained weights of tokens by two models]{
    \begin{minipage}[b]{0.560\textwidth}
    \includegraphics[width=1\textwidth]{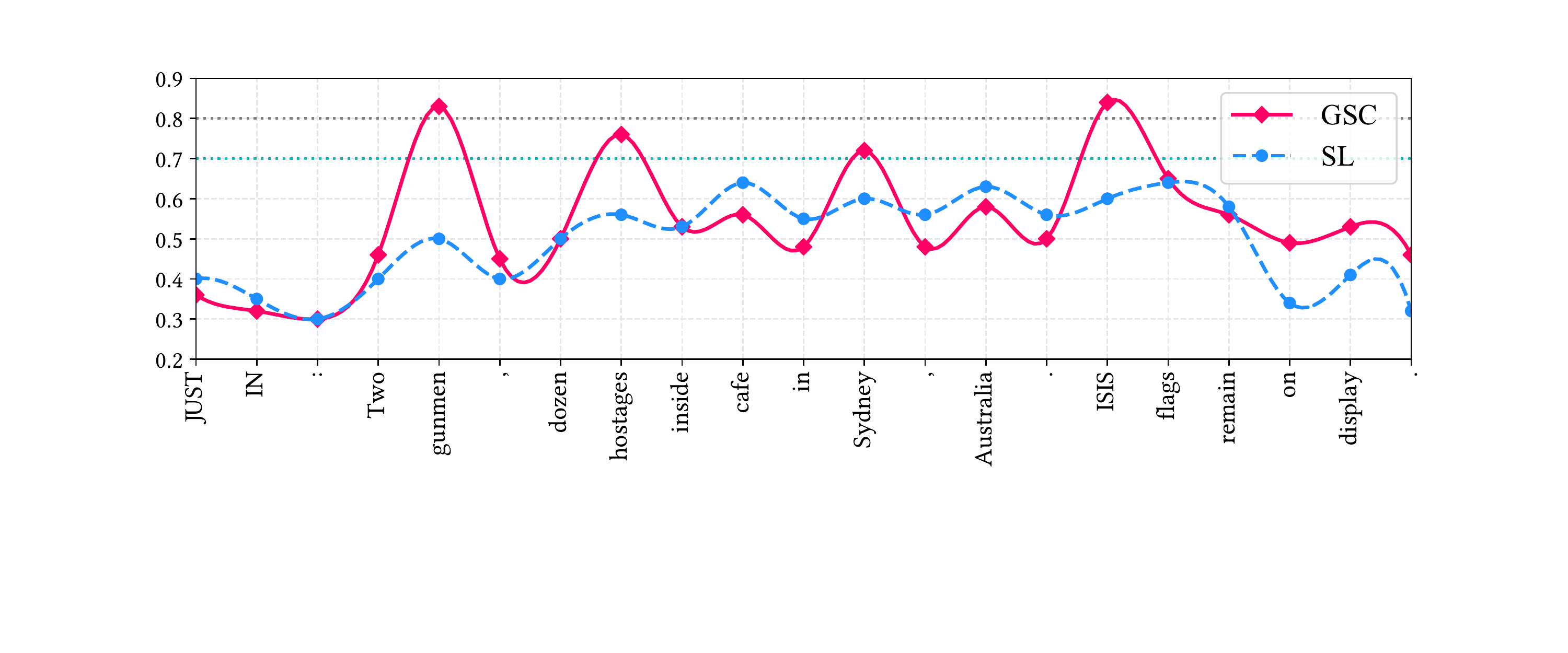}
    \end{minipage}
}
\subfigure[Neuron behaviours of ${\rm \textbf{A}}_{fake}$ and ${\rm \textbf{A}}_{stance}$]{
    \begin{minipage}[b]{0.34\textwidth}
    \includegraphics[width=1\textwidth]{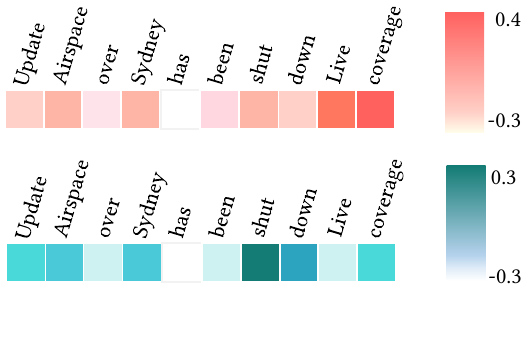}
    \end{minipage}
}
\caption{(a) In fake news detection task, the GSC line denotes the weight values ${\rm \textbf{g}}_{fake}$ of gated sharing cell, while the SL line represents feature weights of ${\rm \textbf{H}}_{shared}$ in the shared layer. Two horizontal lines give two different borders to determine the importance of tokens. (b) The red and green heatmaps describe the neuron behaviours of attention sharing cell ${\rm \textbf{A}}_{fake}$ in fake news detection task and ${\rm \textbf{A}}_{stance}$ in stance detection task, respectively.} \label{Fig5}
\end{figure*}
\subsection{Case Study}

In order to obtain deeper insights and detailed interpretability about the effectiveness of the selected shared layer of the sifted multi-task learning method, we devise experiments to explore some ideas in depth: 1) Aiming at different tasks, what effective features can the selected sharing layer in our method obtain? 2) In the selected sharing layer, what features are learned from different cells?

\subsubsection{The Visualization of Shared Features Learned from Two Tasks}

We visualize shared features learned from the tasks of fake news detection and stance detection. Specifically, we first look up these elements with the largest values from the outputs of the shared layer and the selected shared layer respectively. Then, these elements are mapped into the corresponding values in input embeddings so that we can find out specific tokens. The experimental results are shown in Figure \ref{Fig4}. We draw the following observations:

\begin{itemize}
    \item Comparing PL-FND and PL-SD, private features in private layer from different tasks are different. From PL-FND, PL-SD, and SLT, the combination of the private features and shared features from shared layer increase the diversity of features and help to promote the performance of both fake news detection and stance detection.
    \item By compared SL, SSL-FND, and SSL-SD, selected sharing layers from different tasks can not only filter tokens from shared layer (for instance, `what', `scary', and `fact' present in SL but not in SSL-SD), but also capture helpful tokens for its own task (like `false' and `real' in SSL-FND, and `confirm' and `misleading' in SSL-SD).
\end{itemize}

\subsubsection{The Visualization of Different Features Learned from Different Cells}

To answer the second question, we examine the neuron behaviours of gated sharing cell and attention sharing cell in the selected sharing layer, respectively. More concretely, taking the task of fake news detection as an example, we visualize feature weights of ${\rm \textbf{H}}_{shared}$ in the shared layer and show the weight values ${\rm \textbf{g}}_{fake}$ in gated sharing cell. By that we can find what kinds of features are discarded as interference, as shown in Figure \ref{Fig5}(a). In addition, for attention sharing cell, we visualize which tokens are concerned in attention sharing cell, as shown in Figure \ref{Fig5}(b). From Figure \ref{Fig5}(a) and \ref{Fig5}(b), we obtain the following observations:

\begin{itemize}
    \item In Figure \ref{Fig5}(a), only the tokens ``gunmen, hostages, Sydney, ISIS" give more attention compared with vanilla shared-private model (SP-M). In more details, `gunmen' and `ISIS' obtain the highest weights. These illustrate that gated sharing cell can effectively capture key tokens.
    \item In Figure \ref{Fig5}(b), ``live coverage", as a prominent credibility indicator, wins more concerns in the task of fake news detection than other tokens. By contrast, when the sentence of Figure \ref{Fig5}(b) is applied to the task of stance detection, the tokens ``shut down" obtain the maximum weight, instead of ``live coverage". These may reveal that attention sharing cell focuses on different helpful features from the shared layer for different tasks.
\end{itemize}

\section{Conclusion}

In this paper, we explored a sifted multi-task learning method with a novel selected sharing structure for fake news detection. The selected sharing structure fused single gate mechanism for filtering useless shared features and attention mechanism for paying close attention to features that were helpful to target tasks. We demonstrated the effectiveness of the proposed method on two public, challenging datasets and further illustrated by visualization experiments. There are several important directions remain for future research: (1) the fusion mechanism of private and shared features; (2) How to represent meta-data of fake news better to integrate into inputs.

\section*{Acknowledgments}{The research work is supported by ``the World-Class Universities(Disciplines) and the Characteristic Development Guidance Funds for the Central Universities"(PY3A022), Shenzhen Science and Technology Project(JCYJ20180306170836595), the National Natural Science Fund of China (No.F020807), Ministry of Education Fund Project ``Cloud Number Integration Science and Education Innovation" (No.2017B00030), Basic Scientific Research Operating Expenses of Central Universities (No.ZDYF2017006).}

\bibliographystyle{acl_natbib}
\bibliography{paper6-emnlp-ijcnlp-2019}

\begin{thebibliography}{42}
\expandafter\ifx\csname natexlab\endcsname\relax\def\natexlab#1{#1}\fi

\bibitem[{Allcott and Gentzkow(2017)}]{allcott2017social}
Hunt Allcott and Matthew Gentzkow. 2017.
\newblock Social media and fake news in the 2016 election.
\newblock \emph{Journal of economic perspectives}, 31(2):211--36.

\bibitem[{Castillo et~al.(2011)Castillo, Mendoza, and
  Poblete}]{castillo2011information}
Carlos Castillo, Marcelo Mendoza, and Barbara Poblete. 2011.
\newblock Information credibility on twitter.
\newblock In \emph{Proceedings of the 20th international conference on World
  wide web}, pages 675--684. ACM.

\bibitem[{Chen et~al.(2018)Chen, Awadallah, Hassan, Wang, and
  Cardie}]{chen2018zero}
Xilun Chen, Ahmed~Hassan Awadallah, Hany Hassan, Wei Wang, and Claire Cardie.
  2018.
\newblock Zero-resource multilingual model transfer: Learning what to share.
\newblock \emph{arXiv preprint arXiv:1810.03552}.

\bibitem[{Chen and Cardie(2018)}]{chen2018multinomial}
Xilun Chen and Claire Cardie. 2018.
\newblock Multinomial adversarial networks for multi-domain text
  classification.
\newblock In \emph{Proceedings of the 2018 Conference of the North American
  Chapter of the Association for Computational Linguistics: Human Language
  Technologies}, pages 1226--1240.

\bibitem[{Chen et~al.(2017)Chen, Liu, and Kao}]{chen2017ikm}
Yi-Chin Chen, Zhao-Yang Liu, and Hung-Yu Kao. 2017.
\newblock Ikm at semeval-2017 task 8: Convolutional neural networks for stance
  detection and rumor verification.
\newblock In \emph{SemEval-2017}, pages 465--469.

\bibitem[{Chung et~al.(2014)Chung, Gulcehre, Cho, and
  Bengio}]{chung2014empirical}
Junyoung Chung, Caglar Gulcehre, Kyunghyun Cho, and Yoshua Bengio. 2014.
\newblock Empirical evaluation of gated recurrent neural networks on sequence
  modeling.
\newblock In \emph{NIPS 2014 Workshop on Deep Learning, December 2014}.

\bibitem[{Conroy et~al.(2015)Conroy, Rubin, and Chen}]{conroy2015automatic}
Niall~J Conroy, Victoria~L Rubin, and Yimin Chen. 2015.
\newblock Automatic deception detection: Methods for finding fake news.
\newblock \emph{Proceedings of the Association for Information Science and
  Technology}, 52(1):1--4.

\bibitem[{Derczynski et~al.(2017)Derczynski, Bontcheva, Liakata, Procter, Hoi,
  and Zubiaga}]{derczynski2017semeval}
Leon Derczynski, Kalina Bontcheva, Maria Liakata, Rob Procter, Geraldine
  Wong~Sak Hoi, and Arkaitz Zubiaga. 2017.
\newblock Semeval-2017 task 8: Rumoureval: Determining rumour veracity and
  support for rumours.
\newblock In \emph{SemEval-2017}, pages 69--76.

\bibitem[{Devlin et~al.(2018)Devlin, Chang, Lee, and
  Toutanova}]{devlin2018bert}
Jacob Devlin, Ming-Wei Chang, Kenton Lee, and Kristina Toutanova. 2018.
\newblock Bert: Pre-training of deep bidirectional transformers for language
  understanding.
\newblock \emph{arXiv preprint arXiv:1810.04805}.

\bibitem[{Dungs et~al.(2018)Dungs, Aker, Fuhr, and Bontcheva}]{dungs2018can}
Sebastian Dungs, Ahmet Aker, Norbert Fuhr, and Kalina Bontcheva. 2018.
\newblock Can rumour stance alone predict veracity?
\newblock In \emph{Proceedings of the 27th International Conference on
  Computational Linguistics}, pages 3360--3370.

\bibitem[{Flintham et~al.(2018)Flintham, Karner, Bachour, Creswick, Gupta, and
  Moran}]{flintham2018falling}
Martin Flintham, Christian Karner, Khaled Bachour, Helen Creswick, Neha Gupta,
  and Stuart Moran. 2018.
\newblock Falling for fake news: investigating the consumption of news via
  social media.
\newblock In \emph{Proceedings of the 2018 CHI Conference on Human Factors in
  Computing Systems}, page 376. ACM.

\bibitem[{Guacho et~al.(2018)Guacho, Abdali, Shah, and
  Papalexakis}]{guacho2018semi}
Gisel~Bastidas Guacho, Sara Abdali, Neil Shah, and Evangelos~E Papalexakis.
  2018.
\newblock Semi-supervised content-based detection of misinformation via tensor
  embeddings.
\newblock In \emph{2018 IEEE/ACM International Conference on Advances in Social
  Networks Analysis and Mining (ASONAM)}, pages 322--325. IEEE.

\bibitem[{Hochreiter and Schmidhuber(1997)}]{hochreiter1997long}
Sepp Hochreiter and J{\"u}rgen Schmidhuber. 1997.
\newblock Long short-term memory.
\newblock \emph{Neural computation}, 9(8):1735--1780.

\bibitem[{Kochkina et~al.(2017)Kochkina, Liakata, and
  Augenstein}]{kochkina2017turing}
Elena Kochkina, Maria Liakata, and Isabelle Augenstein. 2017.
\newblock Turing at semeval-2017 task 8: Sequential approach to rumour stance
  classification with branch-lstm.
\newblock \emph{arXiv preprint arXiv:1704.07221}.

\bibitem[{Kochkina et~al.(2018)Kochkina, Liakata, and
  Zubiaga}]{kochkina2018all}
Elena Kochkina, Maria Liakata, and Arkaitz Zubiaga. 2018.
\newblock All-in-one: Multi-task learning for rumour verification.
\newblock \emph{arXiv preprint arXiv:1806.03713}.

\bibitem[{Li et~al.(2018)Li, Zhao, Cheng, and Yang}]{li2018end}
Sizhen Li, Shuai Zhao, Bo~Cheng, and Hao Yang. 2018.
\newblock An end-to-end multi-task learning model for fact checking.
\newblock \emph{EMNLP 2018}, page 138.

\bibitem[{Liu et~al.(2019)Liu, Tan, and Zhou}]{liu2019augmented}
Boyang Liu, Pang-Ning Tan, and Jiayu Zhou. 2019.
\newblock Augmented multi-task learning by optimal transport.
\newblock In \emph{Proceedings of the 2019 SIAM International Conference on
  Data Mining}, pages 19--27. SIAM.

\bibitem[{Liu et~al.(2017)Liu, Qiu, and Huang}]{liu2017adversarial}
Pengfei Liu, Xipeng Qiu, and Xuanjing Huang. 2017.
\newblock Adversarial multi-task learning for text classification.
\newblock In \emph{Proceedings of the 55th Annual Meeting of the Association
  for Computational Linguistics}, pages 1--10.

\bibitem[{Long et~al.(2017)Long, Lu, Xiang, Li, and Huang}]{long2017fake}
Yunfei Long, Qin Lu, Rong Xiang, Minglei Li, and Chu-Ren Huang. 2017.
\newblock Fake news detection through multi-perspective speaker profiles.
\newblock In \emph{Proceedings of the Eighth International Joint Conference on
  Natural Language Processing}, pages 252--256.

\bibitem[{Lukasik et~al.(2016)Lukasik, Srijith, Vu, Bontcheva, Zubiaga, and
  Cohn}]{lukasik2016hawkes}
Michal Lukasik, PK~Srijith, Duy Vu, Kalina Bontcheva, Arkaitz Zubiaga, and
  Trevor Cohn. 2016.
\newblock Hawkes processes for continuous time sequence classification: an
  application to rumour stance classification in twitter.
\newblock In \emph{Proceedings of the 54th Annual Meeting of the Association
  for Computational Linguistics}, volume~2, pages 393--398.

\bibitem[{Ma et~al.(2018{\natexlab{a}})Ma, Gao, and Wong}]{ma2018detect}
Jing Ma, Wei Gao, and Kam-Fai Wong. 2018{\natexlab{a}}.
\newblock Detect rumor and stance jointly by neural multi-task learning.
\newblock In \emph{Companion of the The Web Conference 2018 on The Web
  Conference 2018}, pages 585--593. International World Wide Web Conferences
  Steering Committee.

\bibitem[{Ma et~al.(2018{\natexlab{b}})Ma, Gao, and Wong}]{ma2018rumor}
Jing Ma, Wei Gao, and Kam-Fai Wong. 2018{\natexlab{b}}.
\newblock Rumor detection on twitter with tree-structured recursive neural
  networks.
\newblock In \emph{ACL}, pages 1980--1989.

\bibitem[{Mendoza et~al.(2010)Mendoza, Poblete, and
  Castillo}]{mendoza2010twitter}
Marcelo Mendoza, Barbara Poblete, and Carlos Castillo. 2010.
\newblock Twitter under crisis: Can we trust what we rt?
\newblock In \emph{Proceedings of the first workshop on social media
  analytics}, pages 71--79. ACM.

\bibitem[{Mikolov et~al.(2013)Mikolov, Sutskever, Chen, Corrado, and
  Dean}]{mikolov2013distributed}
Tomas Mikolov, Ilya Sutskever, Kai Chen, Greg~S Corrado, and Jeff Dean. 2013.
\newblock Distributed representations of words and phrases and their
  compositionality.
\newblock In \emph{Advances in neural information processing systems}, pages
  3111--3119.

\bibitem[{Mohtarami et~al.(2018)Mohtarami, Baly, Glass, Nakov, M{\`a}rquez, and
  Moschitti}]{mohtarami2018automatic}
Mitra Mohtarami, Ramy Baly, James Glass, Preslav Nakov, Llu{\'\i}s M{\`a}rquez,
  and Alessandro Moschitti. 2018.
\newblock Automatic stance detection using end-to-end memory networks.
\newblock In \emph{Proceedings of the 2018 Conference of the North American
  Chapter of the Association for Computational Linguistics: Human Language
  Technologies}, volume~1, pages 767--776.

\bibitem[{Mou et~al.(2016)Mou, Men, Li, Xu, Zhang, Yan, and
  Jin}]{mou2016natural}
Lili Mou, Rui Men, Ge~Li, Yan Xu, Lu~Zhang, Rui Yan, and Zhi Jin. 2016.
\newblock Natural language inference by tree-based convolution and heuristic
  matching.
\newblock In \emph{ACL}, page 130.

\bibitem[{Oshikawa et~al.(2018)Oshikawa, Qian, and Wang}]{oshikawa2018survey}
Ray Oshikawa, Jing Qian, and William~Yang Wang. 2018.
\newblock A survey on natural language processing for fake news detection.
\newblock \emph{arXiv preprint arXiv:1811.00770}.

\bibitem[{Popat et~al.(2018)Popat, Mukherjee, Yates, and
  Weikum}]{popat2018declare}
Kashyap Popat, Subhabrata Mukherjee, Andrew Yates, and Gerhard Weikum. 2018.
\newblock Declare: Debunking fake news and false claims using evidence-aware
  deep learning.
\newblock In \emph{EMNLP}, pages 22--32.

\bibitem[{Potthast et~al.(2017)Potthast, Kiesel, Reinartz, Bevendorff, and
  Stein}]{potthast2017stylometric}
Martin Potthast, Johannes Kiesel, Kevin Reinartz, Janek Bevendorff, and Benno
  Stein. 2017.
\newblock A stylometric inquiry into hyperpartisan and fake news.
\newblock \emph{arXiv preprint arXiv:1702.05638}.

\bibitem[{Qian et~al.(2018)Qian, Gong, Sharma, and Liu}]{qian2018neural}
Feng Qian, Chengyue Gong, Karishma Sharma, and Yan Liu. 2018.
\newblock Neural user response generator: Fake news detection with collective
  user intelligence.
\newblock In \emph{IJCAI}, pages 3834--3840.

\bibitem[{Ruchansky et~al.(2017)Ruchansky, Seo, and Liu}]{ruchansky2017csi}
Natali Ruchansky, Sungyong Seo, and Yan Liu. 2017.
\newblock Csi: A hybrid deep model for fake news detection.
\newblock In \emph{Proceedings of the 2017 ACM on Conference on Information and
  Knowledge Management}, pages 797--806. ACM.

\bibitem[{Srivastava et~al.(2015)Srivastava, Greff, and
  Schmidhuber}]{srivastava2015highway}
Rupesh~Kumar Srivastava, Klaus Greff, and J{\"u}rgen Schmidhuber. 2015.
\newblock Highway networks.
\newblock \emph{arXiv preprint arXiv:1505.00387}.

\bibitem[{Thorne et~al.(2017)Thorne, Chen, Myrianthous, Pu, Wang, and
  Vlachos}]{thorne2017fake}
James Thorne, Mingjie Chen, Giorgos Myrianthous, Jiashu Pu, Xiaoxuan Wang, and
  Andreas Vlachos. 2017.
\newblock Fake news stance detection using stacked ensemble of classifiers.
\newblock In \emph{Proceedings of the 2017 EMNLP Workshop: Natural Language
  Processing meets Journalism}, pages 80--83.

\bibitem[{Vaswani et~al.(2017)Vaswani, Shazeer, Parmar, Uszkoreit, Jones,
  Gomez, Kaiser, and Polosukhin}]{vaswani2017attention}
Ashish Vaswani, Noam Shazeer, Niki Parmar, Jakob Uszkoreit, Llion Jones,
  Aidan~N Gomez, {\L}ukasz Kaiser, and Illia Polosukhin. 2017.
\newblock Attention is all you need.
\newblock In \emph{Advances in neural information processing systems}, pages
  5998--6008.

\bibitem[{Wang(2017)}]{wang2017liar}
William~Yang Wang. 2017.
\newblock " liar, liar pants on fire": A new benchmark dataset for fake news
  detection.
\newblock \emph{arXiv preprint arXiv:1705.00648}.

\bibitem[{Wang et~al.(2015)Wang, Wang, Tang, Liu, and
  Li}]{wang2015unsupervised}
Yilin Wang, Suhang Wang, Jiliang Tang, Huan Liu, and Baoxin Li. 2015.
\newblock Unsupervised sentiment analysis for social media images.
\newblock In \emph{Twenty-Fourth International Joint Conference on Artificial
  Intelligence}.

\bibitem[{Wu et~al.(2019)Wu, Rao, Yu, Wang, and Ambreen}]{wu2019multi}
Lianwei Wu, Yuan Rao, Hualei Yu, Yiming Wang, and Nazir Ambreen. 2019.
\newblock A multi-semantics classification method based on deep learning for
  incredible messages on social media.
\newblock \emph{Chinese Journal of Electronics}, 28(4):754--763.

\bibitem[{Wu et~al.(2018)Wu, Rao, Yu, Wang, and Nazir}]{wu2018false}
Lianwei Wu, Yuan Rao, Hualei Yu, Yiming Wang, and Ambreen Nazir. 2018.
\newblock False information detection on social media via a hybrid deep model.
\newblock In \emph{International Conference on Social Informatics}, pages
  323--333. Springer.

\bibitem[{Yang et~al.(2012)Yang, Liu, Yu, and Yang}]{yang2012automatic}
Fan Yang, Yang Liu, Xiaohui Yu, and Min Yang. 2012.
\newblock Automatic detection of rumor on sina weibo.
\newblock In \emph{Proceedings of the ACM SIGKDD Workshop on Mining Data
  Semantics}, page~13. ACM.

\bibitem[{Zhang et~al.(2019)Zhang, Lipani, Liang, and Yilmaz}]{zhang2019reply}
Qiang Zhang, Aldo Lipani, Shangsong Liang, and Emine Yilmaz. 2019.
\newblock Reply-aided detection of misinformation via bayesian deep learning.
\newblock In \emph{Companion Proceedings of The Web Conference}.

\bibitem[{Zubiaga et~al.(2016{\natexlab{a}})Zubiaga, Kochkina, Liakata,
  Procter, and Lukasik}]{zubiaga2016stance}
Arkaitz Zubiaga, Elena Kochkina, Maria Liakata, Rob Procter, and Michal
  Lukasik. 2016{\natexlab{a}}.
\newblock Stance classification in rumours as a sequential task exploiting the
  tree structure of social media conversations.
\newblock In \emph{Proceedings of COLING 2016, the 26th International
  Conference on Computational Linguistics: Technical Papers}, pages 2438--2448.

\bibitem[{Zubiaga et~al.(2016{\natexlab{b}})Zubiaga, Liakata, Procter, Hoi, and
  Tolmie}]{zubiaga2016analysing}
Arkaitz Zubiaga, Maria Liakata, Rob Procter, Geraldine Wong~Sak Hoi, and Peter
  Tolmie. 2016{\natexlab{b}}.
\newblock Analysing how people orient to and spread rumours in social media by
  looking at conversational threads.
\newblock \emph{PloS one}, 11(3):e0150989.

\end{thebibliography}



\end{document}